\documentclass[
]{ceurart}

\usepackage{listings}
\usepackage{inputenc}
\begin{document}

\copyrightyear{2022}
\copyrightclause{Copyright for this paper by its authors.
  Use permitted under Creative Commons License Attribution 4.0
  International (CC BY 4.0).}

\conference{MediaEval'22: Multimedia Evaluation Workshop,
  January 13--15, 2023, Bergen, Norway and Online}

\title{Baseline Method for the Sport Task of MediaEval 2022 with 3D CNNs using Attention Mechanisms}
\renewcommand{\shorttitle}{Sport Task}

\author{Pierre-Etienne Martin}[%
    orcid=0000-0002-9593-4580,
    email=pierre_etienne_martin@eva.mpg.de,
    url=www.eva.mpg.de/comparative-cultural-psychology/staff/pierre-etienne-martin,
]




\address{CCP Department, Max Planck Institute for Evolutionary Anthropology, D-04103 Leipzig, Germany}


\begin{abstract}
This paper presents the baseline method proposed for the Sports Video task part of the MediaEval 2022 benchmark. This task proposes two subtasks: stroke classification from trimmed videos, and stroke detection from untrimmed videos. This baseline addresses both subtasks.
We propose two types of 3D-CNN architectures to solve the two subtasks. Both 3D-CNNs use Spatio-temporal convolutions and attention mechanisms. The architectures and the training process are tailored to solve the addressed subtask.
This baseline method is shared publicly online to help the participants in their investigation and alleviate eventually some aspects of the task such as video processing, training method, evaluation and submission routine.
The baseline method reaches 86.4\% of accuracy with our v2 model for the classification subtask. For the detection subtask, the baseline reaches a mAP of 0.131 and IoU of 0.515 with our v1 model.
\end{abstract}

\maketitle

\section{Introduction}
\label{sec:intro}

Action classification from videos is a popular topic in the computer vision field~\cite{Dataset:UCF101:2012,Dataset:AVA:2018,Dataset:AVA_Kinetics:2020,DBLP:conf/nips/PiergiovanniR20}. In order to solve such task, 2D CNNs were first introduced~\cite{Hakan:2018, NN:SimonyanTwoStream:2014}. Then, to better capture the temporal information from videos, 3D convolution methods emerged~\citep{NN:3DCNN_first:2007,NN:3DCNN:2017}. Optical flow computed from the RGB stream was also investigated in order to boost performance and translate RGB changes into movement information~\cite{NN:I3DCarreira:2017,PeCBMI:2018}. Recently, multi-model methods are re-investigated but this time combining the RGB and the audio streams~\cite{zellers2022merlotreserve} leading to the state-of-the-art on common benchmark datasets such as Kintetics600~\cite{Dataset:Kinetics600:2018}. Multi-view methods combined with Transformers~\cite{dehghani2021scenic} are also the current state-of-the-art in many action classification dataset~\cite{Dataset:EpicKitchens,Dataset:MomentsInTime:2020}.
\par
In the Sport Task of MediaEval 2022, the focus is on the classification and detection of table tennis strokes from videos. As described in~\cite{mediaeval/Martin/2022/overview}, the task focuses on low visual inter-class variability actions: classify them from trimmed videos (subtask 1) and detect them from untrimmed videos (subtask 2). The task is based on TTStroke-21 dataset~\cite{PeMTAP:2020} and is similar to other datasets with low inter-class variability~\cite{Dataset:Gym:2020,Dataset:Diving48:2018,Dataset:EpicKitchens,Mocap:2013}.

This baseline, publicly available on GitHub\footnote{\url{https://github.com/ccp-eva/SportTaskME22}}, tackles the two subtasks and aims to help participants in their submission such as the processing of the videos, the annotation files and the deep learning methods.





\section{Method}
\label{sec:dataset}

The method has been kept simple and uses only the RGB information from the provided videos. The implementation is inspired from~\cite{PeICPR:2020}. The main divergence is the absence of Region Of Interest (ROI) which was computed from Optical Flow values. The data processing is trivial: The RGB frames are resized to a width of 320 and stacked together to form tensors of length 96 either from the trimmed videos or following the annotation boundaries available in the XML files. Data are augmented to increase variability: start at different time points and spatial transformations (flip and rotation).


\par
\begin{figure}
    \centering
    \includegraphics[width=\linewidth]{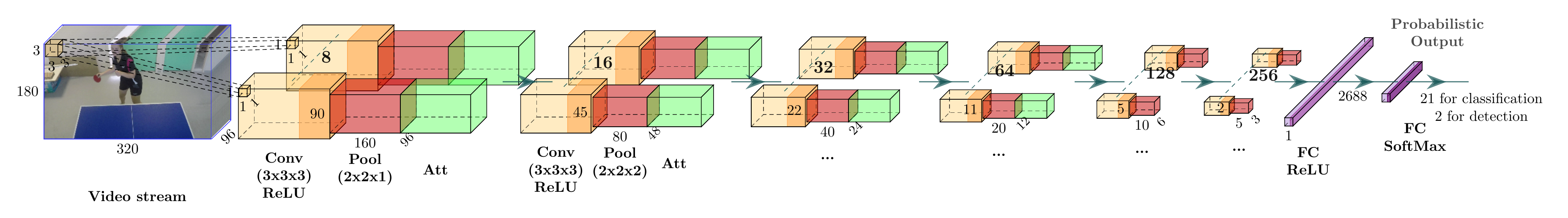}\\
    \includegraphics[width=.91\linewidth]{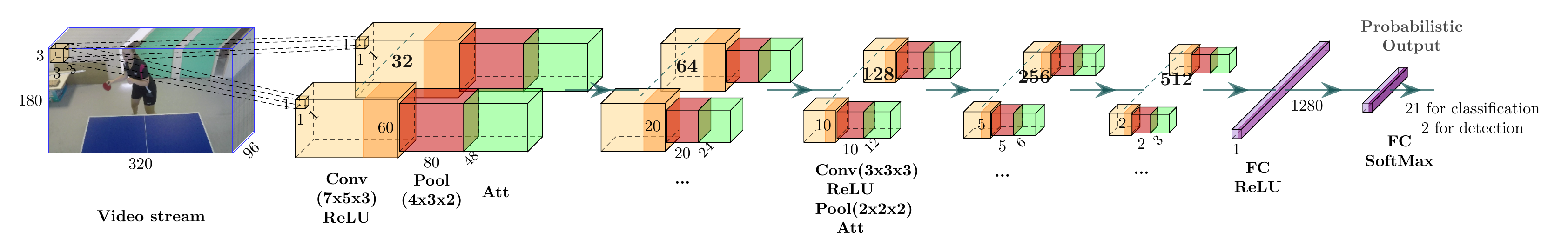}\\
    \caption{3D CNNs v1 (top) and v2 (bottom) using Attention Mechanisms for Stroke Classification and Detection.}
  \label{fig:architecture}
\end{figure}

Two versions, V1 and V2 are introduced and depicted in \autoref{fig:architecture}. V1 is a sequence of four conv+pool+attention layers and two conv+pool layers. All convolutional layers use 3x3x3 filters. The first layers use 2x2x1 pooling filters (no pooling on the temporal domain) and 2x2x2 pooling filters for the other layers. V2 is a sequence of five conv+pool+attention layers. Conv. filters are of size 7x5x3 and pooling filters of size 4x3x2 for the first two blocks. The remaining blocks use 3x3x3 and 2x2x2 for conv. and pooling filters respectively. V2 leads to almost squared feature maps after the second block so that horizontality, verticality and temporality can be better combined before the fully connected layers.

\par

The training method uses Nesterov momentum over a fixed amount of epochs. The learning rate is modified according to the loss evolution~\cite{PeICPR:2020}. The model with the best performance on the validation loss is saved. The training methods are the same for both subtasks. The objective function is the cross-entropy loss of the output processed by the softmax function summing over the batch:
\begin{equation}
    \label{eq:CrossEntropyLoss}
    \mathcal{L}(y,class) = -log(\dfrac{exp(y_{class}')}{\sum_i^Nexp(y_i)})
\end{equation}

\par
We consider 21 classes for the classification task and two classes for the detection task as previously done in~\cite{mediaeval/Martin21/baseline}. Negative samples are extracted for the detection task and negative proposals are built on its test set. Testing is performed with the trimmed proposal (with one window centered or with a sliding window and several post-processing approaches) or by running a sliding window on the whole video for the detection task. The latest output is processed in order to segment framewisely the strokes. Too short strokes, less than 30~frames, are not considered. The model trained on the classification task is also tested on the detection task without further training on the detection data. Two approaches are considered: 1)~Negative class score VS all others for decision and 2)~Negative class score VS sum of all the others. Several decision methods are also tested: No Window, Vote, Mean, and Gaussian according to a temporal window. See~\cite{PeThesis2020} for further details.

\section{Results}

This section presents the results per subtask according to the metrics presented in~\cite{mediaeval/Martin/2022/overview}. For the two subtasks, we trained the models for 2000~epochs using a learning rate of .0001, a momentum of .5 and a weight decay of .005.


\subsection{Subtask 1 - Stroke Classification}

\begin{figure}
    \centering
    \includegraphics[width=.9\linewidth]{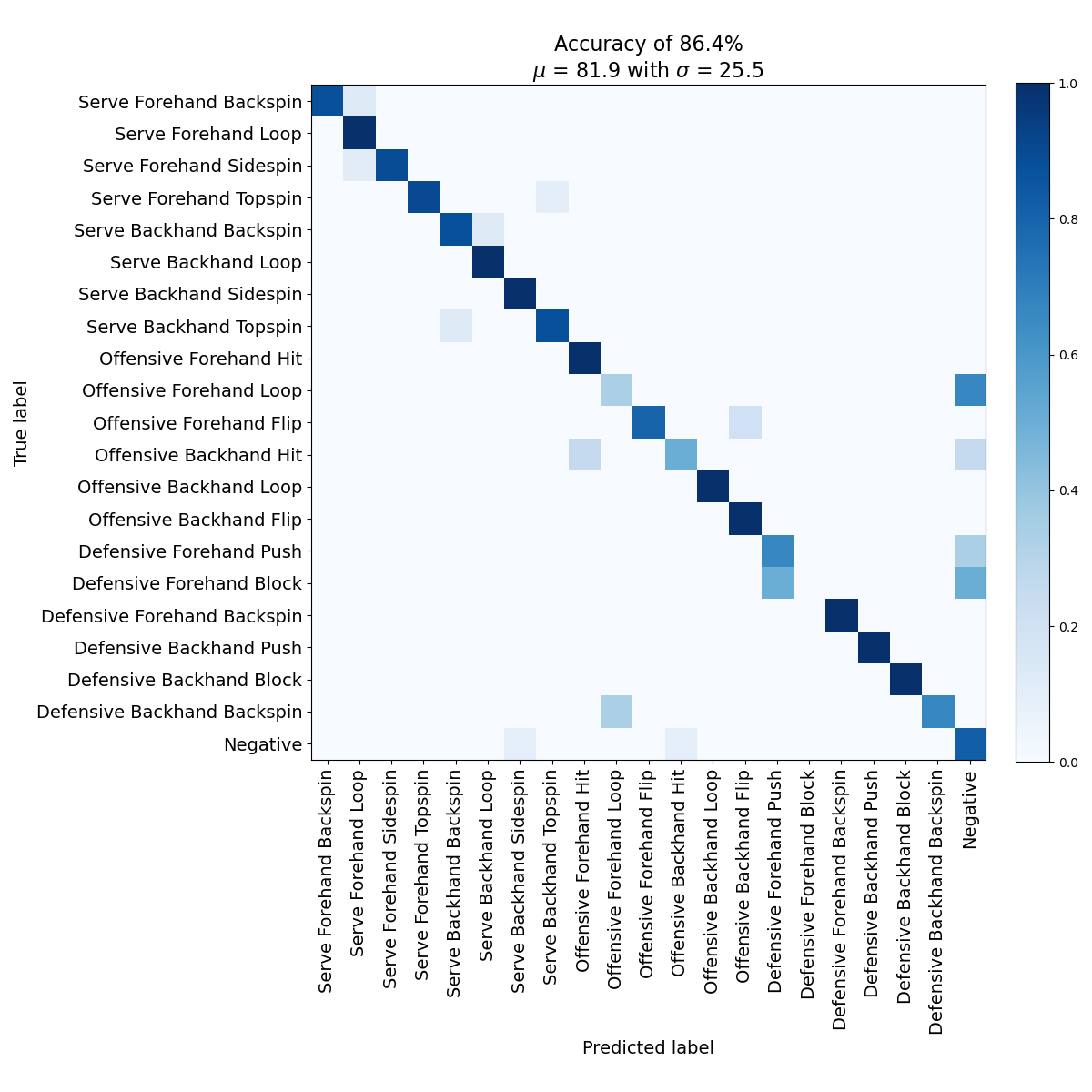}
    \caption{Confusion matrix of the best classification run on the test set.}
    \label{fig:cm}
\end{figure}

As reported in \autoref{Table:Acc}, V1 and V2 perform similarly on the stroke classification subtask, but V2 using the Gaussian window decision performs the best with 86.4\% of accuracy on the test set. This model finished convergence at epoch 815 with train and validation accuracies of .989 and .813 respectively. The confusion matrix of this run is depicted in \autoref{fig:cm}.


\begin{table}
\caption{Models performance on classification subtask in terms of accuracy}
  \label{Table:Acc}
    \begin{tabular}{|c|c|c|c|c|}
    \toprule
        Model & No Window & Vote & Mean & Gaussian \\ \midrule
        V1 & .847 & .839 & .856 & .856 \\ \midrule
        V2 & .856 & .822 & .831 & \textbf{.864} \\ \bottomrule
    \end{tabular}
\end{table}

As we can notice on the confusion matrix, the model has the tendency to classify some strokes as non-strokes (negative class). This is certainly due to the variation in the negative class, increasing its dedicated latent space and giving more probability to the unseen samples to fall in it. This could be solved by increasing the variability of these samples via data augmentation or more recording of these strokes.


\subsection{Subtask 2 - Stroke Detection}

\autoref{Table:Detection1} reports the results using video candidates from the test set. Video candidates are simply non-overlapped successive samples of length 150 frames from the test videos. The main metric for evaluation is the mAP, and therefore the model V2 using a Vote decision performs the best. However, extracting video candidates in such way is not efficient to detect the strokes. That is why in \autoref{Table:Detection2} results using another segmentation methods are reported.

\begin{table}
    \centering
    \caption{Models performance on detection subtask in terms of mAP~|~IoU with proposals on the test set}
    \label{Table:Detection1}
    \begin{tabular}{|c|c|c|c|c|}
        \toprule
        Model & No Window & Vote & Mean & Gaussian \\ \midrule
        V1 & .111 | .358  & .114 | .360 & .113 | \textbf{.365} & .113 | .361 \\ \midrule
        V2 & .111 | .322 & \textbf{.118} | .329 & .117 | .333 & .117 | .331 \\ \bottomrule
    \end{tabular}
\end{table}

To perform a better segmentation, a sliding window with step one is used on the test videos. The outputs are combined in order to make a decision following the same previously presented window methods. The models from subtask 1 are also tested. 

\begin{table}
    \centering
    \caption{Models performance on detection subtask in terms of mAP~|~IoU with sliding window segmentation}
    \label{Table:Detection2}
    \begin{tabular}{|c|c|c|c|}
    \toprule
        Model & Vote & Mean & Gaussian \\ \midrule
        V1 & \textbf{.131} | \textbf{.515} & .00201 | .341 & .00227 | .33 \\ \midrule
        V2 & .000731 | .308 & .102 | .473 & .1 | .466 \\
        \bottomrule
    \end{tabular}
\end{table}

As we can see, the segmentation method allows the model V1 to reach the best performance in terms of mAP and IoU. However it is not the case for the V2 models. 

\section{Conclusion}
\label{sec:discussion}

This baseline intends to help the participants solving the Sports Video Task. This work is in the continuity of last year's baseline~\cite{mediaeval/Martin21/baseline} and more tools were implemented to help the participants as suggested at the last edition. Improvements can be made by combining knowledge from subtask 1 to solve subtask 2. Also, the data augmentation and the loss can be improved to balance the unbalanced distribution of the samples. Finally, the segmentation method for stroke detection can still be improved to boost the performance in this subtask. These possibilities of improvements may be implemented in next year's baseline.

\def\bibfont{\footnotesize} 
\bibliography{references} 

\end{document}